\documentclass[10pt,sigconf,nonacm]{acmart}

\usepackage{graphicx} 
\usepackage{url}            
\usepackage{booktabs}       
\usepackage{amsfonts}       
\usepackage{nicefrac}       
\usepackage{microtype}      
\usepackage{xcolor}         
\usepackage{graphicx}
\usepackage{caption}
\usepackage{subcaption}
\usepackage{algorithm}
\usepackage{algpseudocode}
\usepackage{geometry}
\usepackage{hyperref}
\usepackage{pifont}
\usepackage{soul}
\usepackage{makecell}

\AtBeginDocument{%
  \providecommand\BibTeX{{%
    \normalfont B\kern-0.5em{\scshape i\kern-0.25em b}\kern-0.8em\TeX}}}

\begin{document}

\title{Protea: Client Profiling within Federated Systems using Flower}






\author{%
  Wanru Zhao$^{\dagger}$, Xinchi Qiu$^{\dagger}$, Javier Fernandez-Marques$^{\diamond}$\\ Pedro P. B. de Gusmão$^{\dagger}$, Nicholas D.~Lane$^{\dagger \diamond}$  
  \vspace{+0.1cm}
  }

 	
 \affiliation{\institution{$^\dagger$University of Cambridge\hspace{+0.75cm}$^\diamond$Samsung AI}\country{}}
\renewcommand{\shortauthors}{}

\begin{abstract}
Federated Learning (FL) has emerged as a prospective solution that facilitates the training of a high-performing centralised model without compromising the privacy of users. While successful, FL research is currently limited by the difficulties of establishing a realistic large-scale FL system at the early stages of  experimentation. Simulation can help accelerate this process. To facilitate efficient scalable FL simulation of heterogeneous clients, we design and implement {\tt Protea}, a flexible and lightweight client profiling component within federated systems using the FL framework Flower
. It allows automatically collecting system-level statistics and estimating the resources needed for each client, thus running the simulation in a resource-aware fashion. The results show that our design successfully increases parallelism for 1.66 $\times$ faster wall-clock time and 2.6$\times$ better GPU utilisation, which enables large-scale experiments on heterogeneous clients. 


\end{abstract}

\maketitle

\thispagestyle{fancy}            
\fancyhead{}                     
\chead{PREPRINT: Accepted at ACM MobiCom FedEdge Workshop, 2022}                
 
 

\section{Introduction}


As Federated Learning (FL)~\cite{mcmahan2017communication,bonawitz2019towards} matures, more experimentation is required before deployment in various applications in the real world. However, research on large-scale FL is challenging in terms of scale and systems heterogeneity~\cite{kairouz2019advances}, and simulating such scenarios has become the go-to approach.

Simulation can be highly beneficial if it allows enough degrees of freedom to study different scenarios at the system and algorithmic level. Recently, several frameworks have been proposed to support simulating FL workloads, such as FedML ~\cite{chaoyanghe2020fedml}, FedJAX~\cite{fedjax}, FLUTE~\cite{flute},  PrivacyFL~\cite{PrivacyFL}, OpenFL ~\cite{reina2021openfl} and EasyFL ~\cite{EasyFL}. A typical procedure to launch an FL simulation requires specifying how many clients operate in each round and the resources each client needs. However, few frameworks provide this interface. For example, FedScale~\cite{lai2021fedscale} proposes a resource manager that queues the over-committed simulation tasks of clients and adaptively schedules them at runtime. 


However, knowing how many resources are needed for each client in advance could be tedious, since it has to be repeated for different hyperparameter choices (e.g., batch sizes), or every time a new model is used. To make things worse, it becomes even harder when clients become heterogeneous with different models or different hyperparameters. Furthermore, setting the resources for each client inappropriately could lead to severe consequences: if we unreasonably allocate too few resources to clients, out-of-memory errors will occur; on the other hand, over-allocating resources to a client can lead to under-utilisation of system resources, thus resulting in longer training time and limiting scalability. 

To accurately measure the resources needed for each client, one option is to use existing framework-specific profilers. Every deep learning framework has an API to monitor the statistics of the GPU devices, such as Pytorch Profiler~\cite{paszke2019pytorch}, TensorFlower Profiler~\cite{abadi2016tensorflow} and MXNet Profiler~\cite{chen2015mxnet}. While the profiling tools have become essential to identify computing bottlenecks or memory leaks, they often result in a severe slow down during training, because they extract a comprehensive list of metrics from the workload at hand. Therefore, they are only framed to be run for debugging purposes.

For the purposes of extending the Federated Learning frameworks with a client-side profiler, in this work, we introduce {\tt Protea}, a framework-agnostic and lightweight profiler that enables the collection of metrics about the performance of heterogeneous clients at different levels (e.g. memory and training time). {\tt Protea} can be integrated into Flower~\cite{flower}, a novel end-to-end FL framework, by profiling clients at runtime and updating the system resources allocated to them with the aim of maximising resource utilisation. {\tt Protea} also operates transparently to the user. 

\section{System Design}

This section presents the design goals of {\tt Protea} to address limitations of FL simulation, and describes {\tt Protea} in details.

\subsection{Design Goals}

Many frameworks provide support for simulation, but there are weaknesses in their designs that hinder researchers and developers from using them. Based on these observations, we propose a set of independent design goals for {\tt Protea}:

\textbf{1) Usability}: Given the difficulty in configuring FL simulation, {\tt Protea} should be intuitive and easy to use.

\textbf{2) Flexibility}: Given the rate of change in FL and the speed of the FL systems, {\tt Protea} should provide a flexible API for users to configure parameters for experiment goals. 

\textbf{3) Compatibility}: Given the robust and diverse range of existing ML frameworks, {\tt Protea} should be compatible with multiple deep learning frameworks and general FL systems.

\textbf{4) Efficiency}: Given that real-world FL is often used on large systems, {\tt Protea} should not incur significant communication or computation overhead for FL training.

\textbf{5) Scalability}: Given that real-world FL would encounter a large number of clients, {\tt Protea} should be able to help launch large-scale FL training at an acceptable speed (wall-clock execution time).



A client profiler with those properties will increase both realism and scale in FL research and provide a smooth transition from research in simulation to large-cohort research on real-edge devices. The next section describes how {\tt Protea} supports those goals. 

\subsection{Design of Client Profiling} 

The proposed {\tt Protea} obtains estimates of the GPU and CPU memory utilisation for each client performing local training. The measurements serve as an input to the resource scheduling strategy that allocates hardware resources across active clients in the next round of training. 
\begin{table}[bp]
\centering
  \caption{List of metrics tracked by \texttt{SystemMonitor}, {\tt Protea}'s monitoring thread.}
  \label{tab:metrics}
  \begin{tabular}{cc}
    \toprule
    Metrics & Description \\
    \midrule
    CPU & CPU load of a FL client \\
    RAM & Memory footprint of a FL client\\
    GPU & GPU compute utilisation of a FL client\\
    VRAM & GPU memory usage of a FL client \\
    CPU\_time & CPU time of a FL client \\
    CUDA\_time & GPU time of a FL client \\
  \bottomrule
\end{tabular}
\end{table}

{\tt Protea} introduces  \texttt{SystemMonitor}, a new class derived from \texttt{threading.Thread} that keeps track of CPU, RAM, GPU and VRAM usage (each GPU separately) by pinging for information every 0.7 seconds in a separate thread. We use the locking mechanism to guarantee the synchronisation of threads. Each client profiling thread is bound to the ID of a particular client and runs in parallel to the local training routine. This enables us to specifically address both our \textbf{Flexibility} and \textbf{Efficiency} design goals.

To obtain statistics about CPUs on running processes, we make use of {\tt psutil} (process and system utilities), which is a platform-agnostic package for Python implementing many functions provided by traditional UNIX command line tools.

To fetch statistics about GPUs for a specific process, we use {\tt gputil}, a Python package that provides an interface to the NVIDIA command line tool for their system management interface ({\tt nvidia-smi}).

Table~\ref{tab:metrics} shows the metrics of the \texttt{SystemMonitor} object inside {\tt Protea} tracks using {\tt psutil} and {\tt gputil}.



\section{System Execution in Flower}

Our design is generic to all federated systems, and in this section we illustrate how {\tt Protea} can be integrated into Flower \cite{flower} and improve the resource scheduling mechanism.

\begin{figure}[b]
    \small
    \centering
    \includegraphics[width=0.3\textwidth]{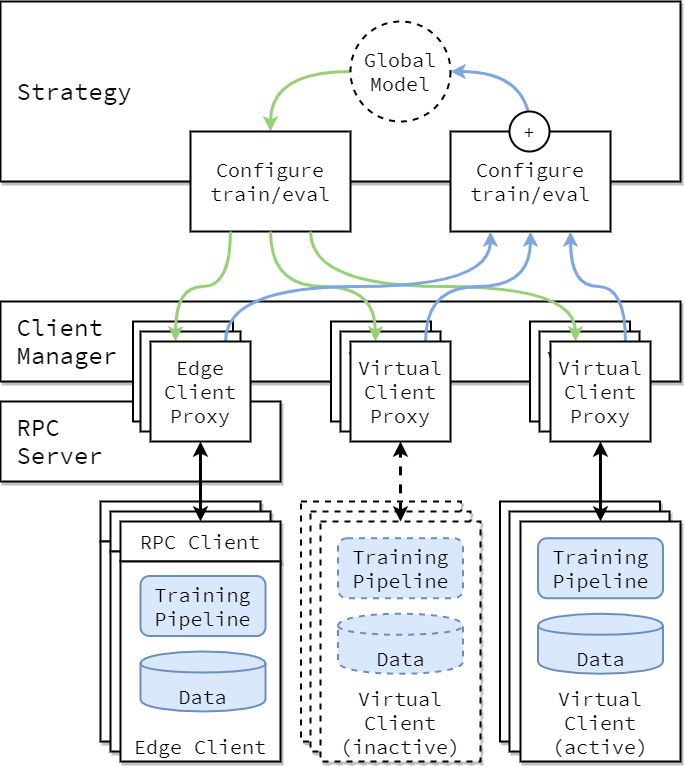}
    \caption{Flower core framework architecture. }
    \vspace{-0.3cm}
    \label{fig:framework-architecture}
\end{figure}

\subsection{Framework Selection – Flower}


Flower (Figure \ref{fig:framework-architecture}) is a comprehensive end-to-end FL framework that distinguishes itself from existing alternatives by providing higher-level abstractions to execute large-scale FL experiments. It also incorporates various heterogeneous FL devices scenarios, which enables a seamless transition from experimental research in simulation to system research on a large cohort of real edge devices. Flower supports different ML frameworks to ensure that FL developers are not constrained by the software stack on client devices. It already supports multiple frameworks including PyTorch~\cite{paszke2019pytorch}, Tensorflow~\cite{abadi2016tensorflow}, MXNet~\cite{chen2015mxnet}, and JAX~\cite{jax2018github}, and extensions to other frameworks such as Flashlight for C++ are ongoing. 

There have been many works being done based on Flower in various fields, including on-device training~\cite{zerofl}, cryptography and security~\cite{secagg}, self-supervised learning~\cite{fedssl} and end-to-end speech recognition~\cite{e2e}. We also note that Flower previously lacked support for profiling, which is a significant limitation during resource scheduling in one round of FL training. 

\subsection{Resource-aware Scheduling}

Flower proposes the concept of virtual clients that are scheduled by the Virtual Client Engine (VCE) over scalable High Performance Computing (HPC) nodes. The VCE is a key module inside the Flower framework enabling running large-scale FL workloads with minimal overhead in a scalable manner. Virtual clients in simulation can also offer a coherent interface with real edge clients to allow for a seamless transition between simulation and deployment. 

\begin{figure}[t]
    \small
    \centering
    \includegraphics[width=0.45\textwidth]{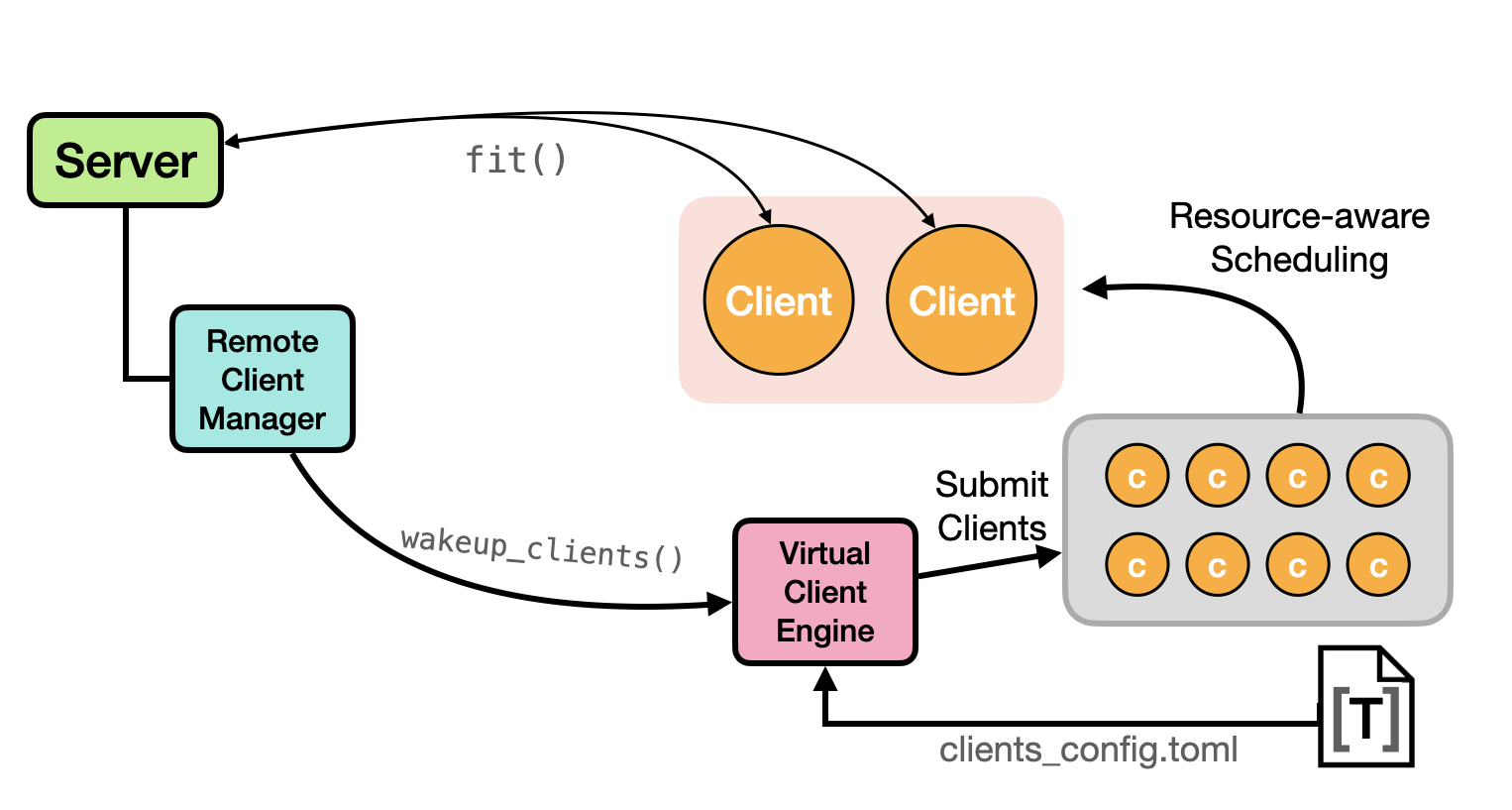}
    \caption{High level diagram flow of a fit round (i.e. on-device training) when Flower operates with clients are simulated.} 
    \vspace{-0.3cm}
    \label{fig:vce}
\end{figure}

The Virtual Client Engine uses Ray~\cite{ray} under the hood to provide an accessible but powerful mechanism to run FL workloads in a resource-aware fashion. At a glance, the Virtual Client Engine operates as follows (shown in Figure~\ref{fig:vce}): (1) first, a user-defined configuration file (e.g. a YAML file) specifying, among other parameters, the resources a client will need (e.g. 2 CPUs and 3.5GB of VRAM) is passed to the server; (2) each client asked to participate in the round will be assigned the resources indicated in the configuration file; (3) Ray will schedule the clients in the round in a FIFO fashion with as many clients running concurrently as the available system resources can hold; (4) once a client has finished training, the resources get freed and another client in the round will be spawned. Stages (3) and (4) are repeated until all clients in the round have participated.

Notably, the resource-aware scheduling mechanism implemented with Ray in Flower requires the user to supply resources for clients. It is not taking into consideration the prospective memory use of a client's workload. Also, asking the user to supply the right client resources would be a complicated process as well.
\hspace{-10mm}
\subsection{Automating Resources Allocation}

\begin{figure}[b]
    \small
    \centering
    \includegraphics[width=0.3\textwidth]{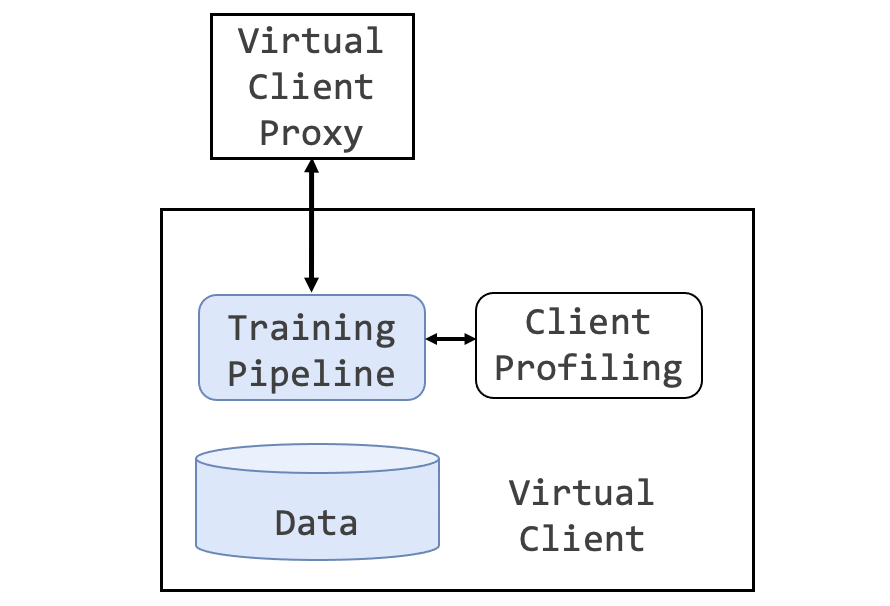}
    \caption{{\tt Protea}'s client-side logic.} 
    \vspace{-0.3cm}
    \label{fig:client-side}
\end{figure}

To update the resources needed for each clients, {\tt Protea} proposes two major components: the server-side logic and the client-side logic (Figure \ref{fig:client-side}).

\textbf{Protea's client-side logic.} \texttt{FlowerClient} implements a new method \texttt{get\_properties()} to enable server-side strategies to query client properties, which returns a dictionary containing client properties (e.g., does this client use a GPU? How much VRAM is the training making use of? What’s the current battery state? How long did it take to do the training last time this client participated? ). 



\textbf{Protea’s server-side logic.} The strategy abstraction in Flower enables the implementation of fully custom strategies. A strategy is the federated learning algorithm that runs on the server deciding how to sample clients, how to configure clients for training, how to aggregate updates, and how to evaluate models. We address our \textbf{Usability} and \textbf{Compatibility} design goals by allowing {\tt Protea} to be compatible with this strategy abstraction. Users can choose to use {\tt Protea} or not easily and flexibly.

We derive a new class from the abstraction that we name {\tt ResourceAwareFedAvg}, which enhances the vanilla Federated Averaging (FedAvg)~\cite{mcmahan2017communication} strategy, where we incorporate properties retrieved from the clients and use them in subsequent rounds to make better use of the available resources. 



In a Flower strategy, method \texttt{configure\_fit()} is responsible for selecting clients and deciding what hyperparameters / configuration to send to each client for training. Before client sampling, we update resources for clients. {\tt Protea} iterates over data from clients in the previous round, finds specific clients in the client manager and updates the resources assigned to them. The results of a fit round are processed with method \texttt{aggregate\_fit()} and aggregates them to generate a new global model. {\tt Protea} enumerates each client in the fit results, gets resources and saves them to be updated in the \texttt{configure\_fit()} next round.

\subsection{Updated Resource Scheduling}

By default, Ray will see all resources in the system (all CPUs, all GPUs, etc.), but this can also be customised when initialising the Virtual Client Engine as well. When specifying GPU resources for each client, Ray requires a \texttt{num\_gpus} to be a ratio in the range of $[0:G]$, where $G$ is the number of GPUs in the system. We introduce a simple helper function that converts the VRAM amount specified by the user in the configuration file into a ratio that can be understood by Ray.

Once the information from {\tt Protea} is obtained, we can calculate using the following formula:
\begin{equation}
{\tt num\_gpus}=\frac{\tt vram\_measured\_for\_single\_worker}{\tt total\_vram\_in\_system}
\end{equation}
Then, it can be passed to simulation and set as the GPU resource for each worker along with other client-specific resources.

\subsection{Better System Resource Utilisation}

The original setup results in sequentially executing each client in the round while {\tt Protea} significantly reduces the wall-clock time by accurately assigning resources when simulating FL workloads. This enables us to specifically address our \textbf{Scalability} design goal. 

\begin{figure*}[ht]
    \centering
    \begin{subfigure}[b]{0.33\textwidth} 
        \centering
        \includegraphics[width=\textwidth]{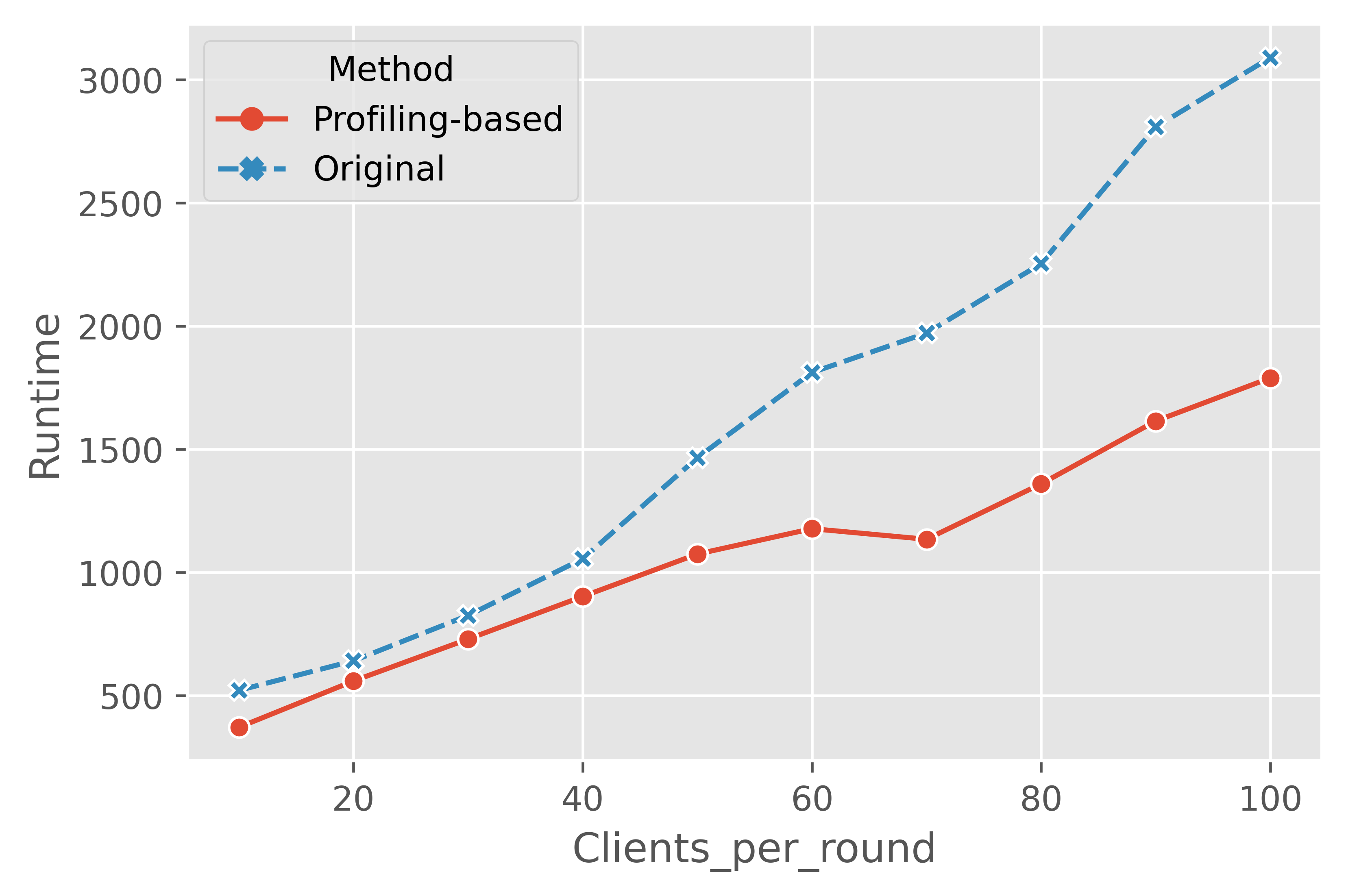}
        \caption{CIFAR-10}
        \label{cifar10}
        \end{subfigure}
    \begin{subfigure}[b]{0.33\textwidth} 
        \centering
        \includegraphics[width=\textwidth]{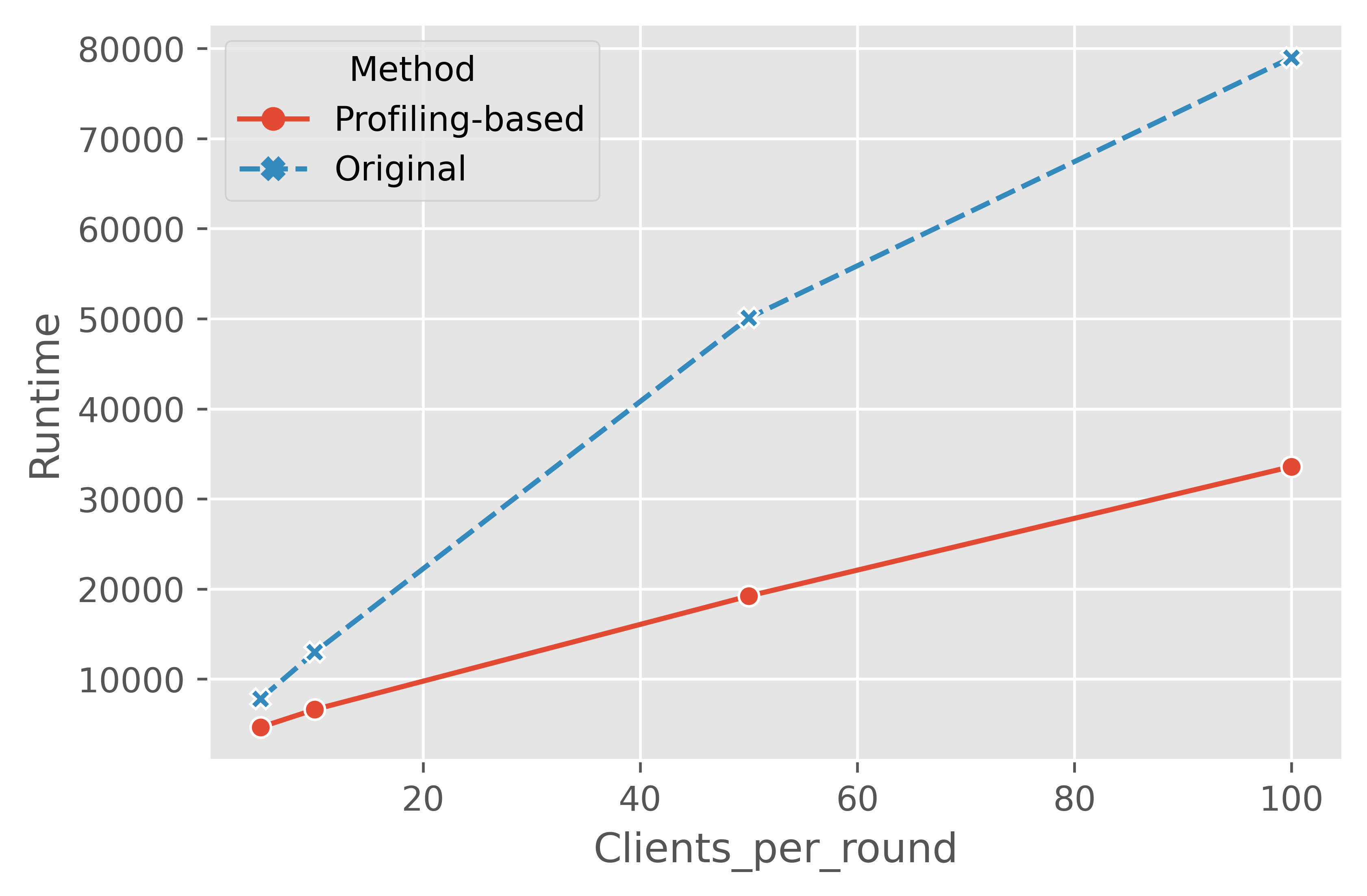}
        \caption{FEMNIST}
        \label{femnist}
        \end{subfigure}
    \begin{subfigure}[b]{0.33\textwidth} 
        \centering
        \includegraphics[width=\textwidth]{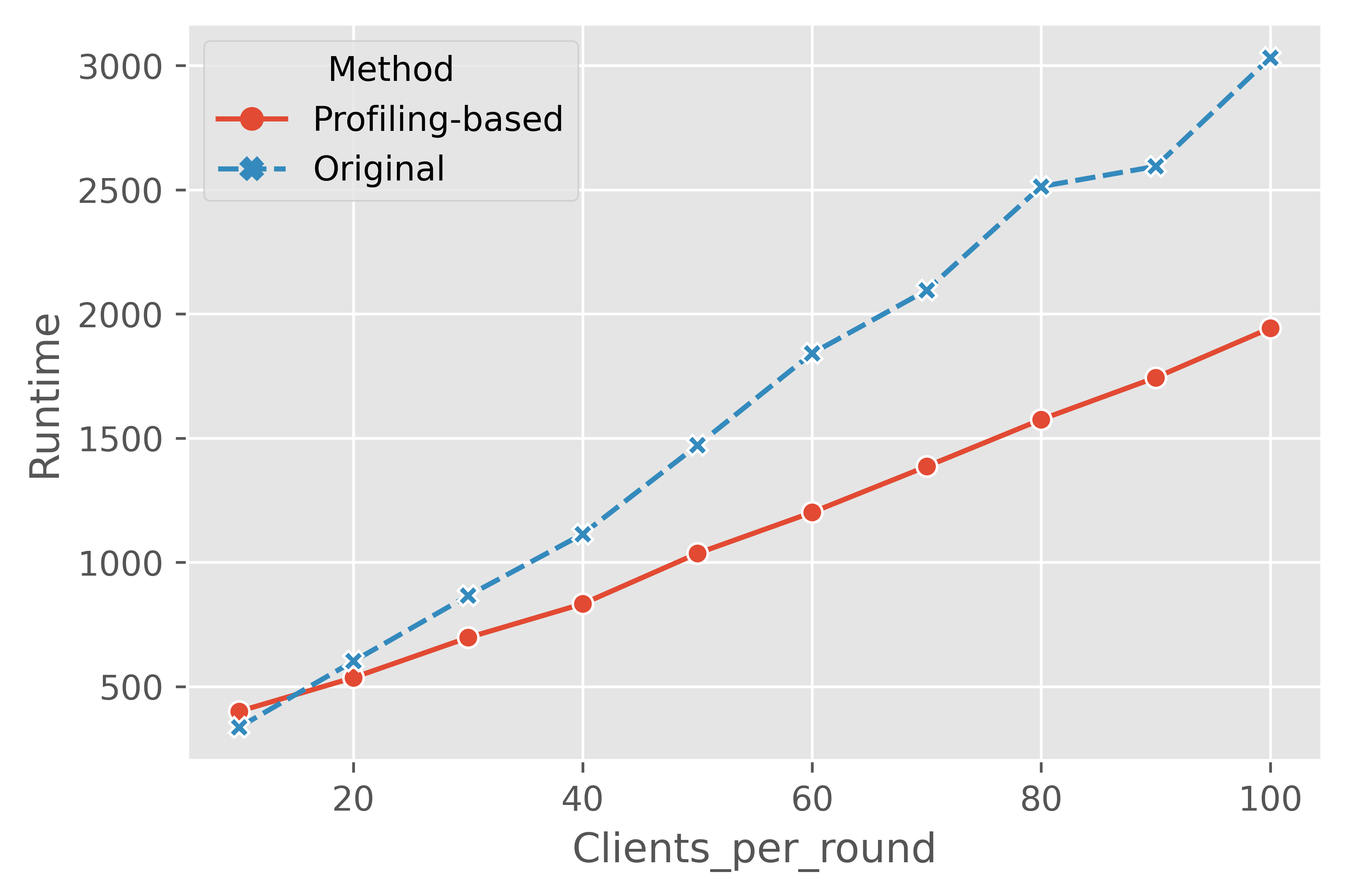}
        \caption{CIFAR-10 with different batch sizes}
        \label{batchsize}
        \end{subfigure}
    \caption{The relationship between the number of clients per round and the runtime (in seconds). A Profiler-enhanced strategy, achieves large wall-clock reductions in FL settings with large cohorts. As more clients participate, assigning the right amount of resources to each client becomes more important. With the proposed Profiler, a Strategy can leverage the collected statistics to more intelligently sample and schedule clients in each round.}
\end{figure*}

\begin{figure}[ht]
  \centering
  \begin{subfigure}[b]{0.235\textwidth}
    \centering
    \includegraphics[width=\textwidth]{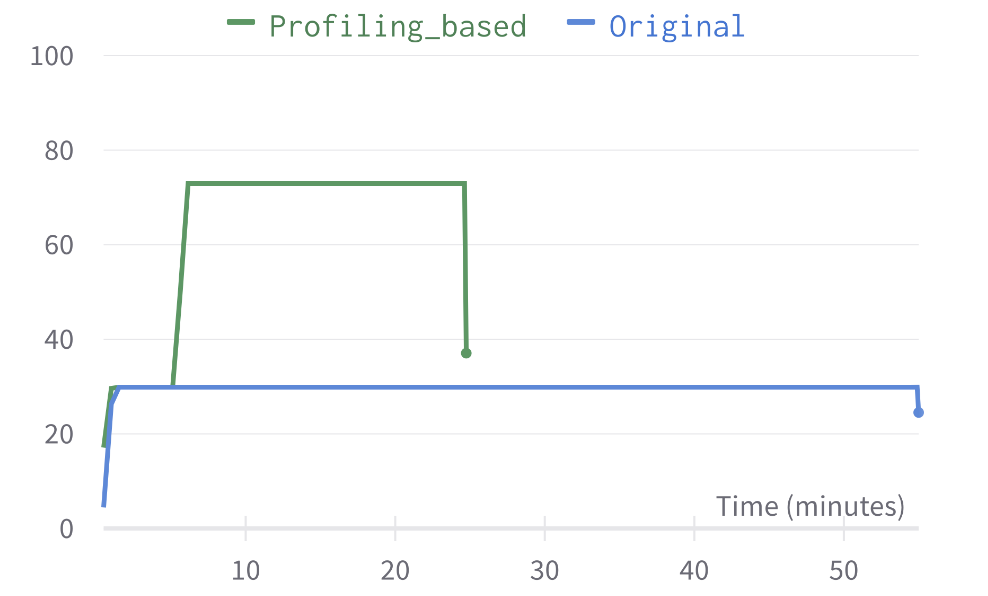}
    \caption{GPU Memory Allocated (\%)} 
    \label{memory}
  \end{subfigure}
  \begin{subfigure}[b]{0.235\textwidth}
    \centering
    \includegraphics[width=\textwidth]{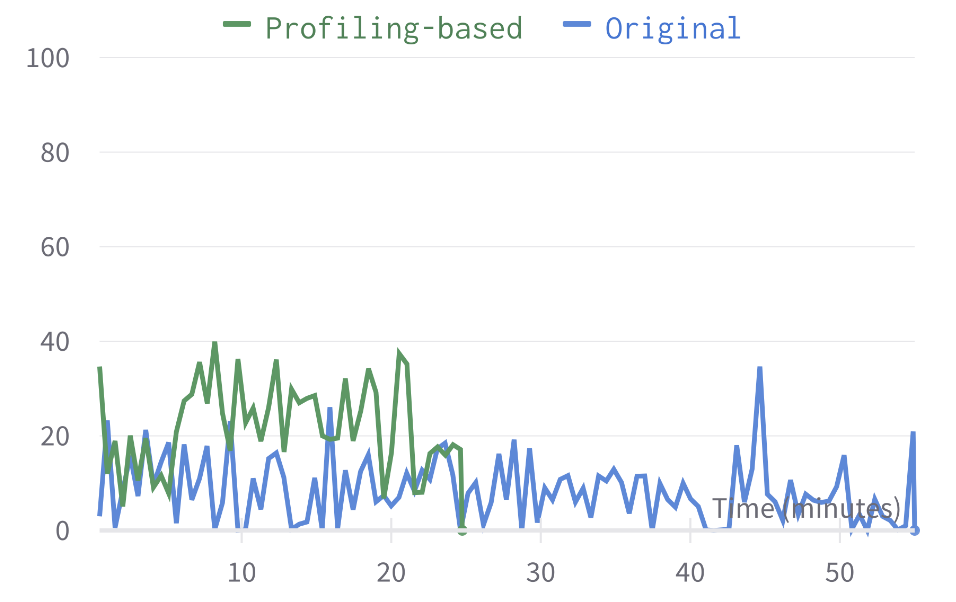}
    \caption{GPU utilisation (\%)} 
    \label{uilization}
  \end{subfigure}
  \captionsetup{font=small,labelfont=bf}
  \caption{The percentage of process GPU memory allocated and utilisation when 100 clients are selected in each round with CIFAR-10. {\tt Protea} achieves much higher GPU utilisation, significantly reducing wall-clock time.}
  \label{cifar10-gpu}
\end{figure}

\section{Evaluation}

This section aims to investigate whether the information retrieved by the {\tt Protea} can be better utilised to support efficient large-scale simulations, and help our resource-aware scheduling to reduce the wall-clock time of training. This is an important evaluation that can shed light on the scalability and heterogeneity of FL training at the system level.
\subsection{Experimental Setup}
Experiments are carried out on two image classification tasks with different complexities in regards to the sample count and the class size: FEMNIST~\cite{leaf} and CIFAR10~\cite{krizhevsky2009learning}, where FEMNIST is built by partitioning the data of the Extended MNIST~\cite{emnist} depending on the digit-character writers. The CIFAR-10 dataset contains 50k and 10k 32$\times$32 RGB images in its training and test sets respectively comprising 10 classes. The FEMNIST dataset, on the other hand, is comprised of 671k 28$\times$28 grey-scale images spanning a total of 62 classes. In both instances, we randomly select 10\% of the training set for validation at the client level, i.e., the validation set for each client is derived from the training partition for each client. It guarantees that the validation set accurately represents the underlying distribution of each client.

Following the convention for CIFAR10, a ResNet-18 \cite{resnet} architecture is implemented, with a \texttt{pool\_size} of 100 and the \texttt{clients\_per\_round} increasing from 10 to 100. For FEMNIST, we employ the much smaller CNN model first proposed in \cite{leaf}. Since FEMNIST is a naturally partitioned dataset by writers' ID, it consists of a \texttt{pool\_size} of 3597, and we vary the  \texttt{clients\_per\_round} from 5 to 3000. More specifically, we increase the number of clients selected in each round to see the trajectory of how wall-clock time and GPU utilisation change. We conducted all the simulation experiments on NVIDIA GeForce GTX 1080Ti GPUs.

\subsection{Experiment I: Homogeneous Clients}
We evaluate the simulation performance with all clients having the same batch size, showing homogeneous resources.

From Figure~\ref{cifar10} and Figure~\ref{femnist}, we can see that the total time of the simulation with our {\tt Protea} integrated into Flower is reduced compared with the original Flower, with 1.56$\times$ and 2.35$\times$ wall clock time reduction respectively for CIFAR-10 dataset and FEMNIST dataset.

From Figure~\ref{cifar10-gpu}, we can see that the GPU utilisation percentage of the simulation with our {\tt Protea} integrated into Flower is 2.6$\times$ greater compared with the original Flower, showing that our strategy can make better use of the underlying resources. 


\subsection{Experiment II: Heterogeneous Clients}


We change the batch size of different clients, which simulates the system heterogeneity, because more VRAM or memory required of a client is equivalent to having a smaller batch size in the setting.

We considered three very different values of {\tt batch\_size}: 32, which is the typical batch size used for CIFAR-10 with ResNet-18; as well as 1024 and 2048, two much larger batch sizes aimed at simulating clients that have larger memory sizes. In the experiment, one third of the clients adopt the batch size of 32, one third of 1024 and another of 2048. 

Figure~\ref{batchsize} shows that our strategy can better deal with heterogeneous resources, with 1.66$\times$ wall clock time reduction.

\section{Implications and Limitations} 

We can observe that our {\tt Protea} can help achieve a significant speedup and better GPU utilisation in settings with homogeneous resources as well as settings with heterogeneous resources, so it can potentially help involve more clients training in each round.

Since simulation can help with testing different scenarios quickly, and {\tt Protea} enables simulation to be more efficient by taking less time and making better use of resources, the profiler would be significantly beneficial to do better federated learning on edge devices in the real world by doing research first on simulation.

However, we recognise a few limitations of our current work and we point to future work to address these issues.

Firstly, memory utilisation might not always correlate with compute utilisation: e.g., some clients might have smaller models but require more operations. We aim to investigate the case of different models in FL scenarios in the future.

Secondly, the use of single-node simulators for large-scale experiments results in excessively long training times, so another very common scenario for researchers is training with multiple GPUs. Our experimental method and conclusions about Flower can be easily extended to this scenario using a cluster of GPUs. Ray guarantees this extendability. However, it cannot support training on heterogeneous hardware at the same time. This is caused by the nature of Ray. We hope to extend our {\tt Protea} to allow users to train their models on clusters with various underlying hardware, including users' own resource types.
\section{Conclusion} 

This paper presents a novel client profiling mechanism {\tt Protea}, which can be fitted within both general FL systems and the recent comprehensive FL framework Flower, to achieve scalable FL workloads on heterogeneous clients. The proposed profiler simplifies the parameterisation of FL workloads since it automates the choice of computing and memory resources needed for each client. The profiler is lightweight and it can run as a parallel process without slowing down the training in a manner transparent to the researchers. Our client profiling component built-in Flower can achieve state-of-the-art performance on both classic and large-scale datasets, i.e., 1.66$\times$ speedup and 2.6$\times$ GPU utilisation on heterogeneous clients, clearly demonstrating its superior usability, flexibility, compatibility, efficiency and scalability.


\bibliographystyle{ACM-Reference-Format}
\bibliography{ref}

\end{document}